%% file: root.tex
\documentclass[letterpaper, 10 pt, journal, twoside]{IEEEtran}

\IEEEoverridecommandlockouts
\usepackage{graphics} %
\usepackage{epsfig} %
\usepackage{times} %
\usepackage{amsmath} %
\usepackage{amssymb}  %
\usepackage{bbm}
\usepackage{booktabs}
\usepackage{mathtools}
\usepackage{multirow}
\usepackage{units}
\usepackage{bm}
\usepackage{xcolor}
\usepackage{hyperref}

\usepackage{enumitem}
\usepackage[ruled,vlined]{algorithm2e}
\usepackage{cite}

\usepackage{letltxmacro}
\LetLtxMacro{\originaleqref}{\eqref}
\renewcommand{\eqref}{Eq.~\originaleqref}

\SetCommentSty{mycommfont} %
\usepackage{cite}
\usepackage{microtype}

\input{tex/math}

\title{Probabilistic Appearance-Invariant\\Topometric Localization with New Place Awareness}

\author{Ming Xu, Tobias Fischer, Niko S{\"u}nderhauf and Michael Milford%
\thanks{Manuscript received: February, 24, 2021; Revised June, 3, 2021; Accepted June, 29, 2021.}%
\thanks{This paper was recommended for publication by Editor Sven Behnke upon evaluation of the Associate Editor and Reviewers' comments.
This work was supported by the Australian Government, via grant AUSMURIB000001 associated with ONR MURI grant N00014-19-1-2571. All authors acknowledge continued support from the Queensland University of Technology (QUT) through the Centre for Robotics.}%
\thanks{The authors are with the QUT Centre for Robotics, Queensland University of Technology, Brisbane, QLD 4000, Australia (e-mail: {\footnotesize \href{mailto:mingda.xu@hdr.qut.edu.au}{mingda.xu@hdr.qut.edu.au}}).}%
\thanks{Digital Object Identifier (DOI): 10.1109/LRA.2021.3096745}%
}
\IEEEpubid{\begin{minipage}{\textwidth}\ \\[12pt] \centering
  \copyright 2021 IEEE. Personal use of this material is permitted.  Permission from IEEE must be obtained for all other uses, in any current \\or future media, including reprinting/republishing this material for advertising or promotional purposes, creating new collective \\works, for resale or redistribution to servers or lists, or reuse of any copyrighted component of this work in other works.
\end{minipage}}

\newcommand{\githublink}{\url{https://github.com/mingu6/TopometricLoc}}

\begin{document}

\bstctlcite{IEEEexample:BSTcontrol}

\maketitle

\input{tex/0-abstract}
\input{tex/1-introduction}
\input{tex/2-relatedworks}

\input{tex/3-methods}
\input{tex/4-experiments}
\input{tex/5-results}
\input{tex/6-conclusion}

\input{./root.bbl}
\end{document}

%% file: tex/math.tex
\newcommand{\bmu}{\bm{\mu}}

\newcommand{\bSigma}{\bm{\Sigma}}
\newcommand{\balpha}{\bm{\alpha}}
\newcommand{\bbeta}{\bm{\beta}}

\newcommand{\bE}{\mathbf{E}}

\newcommand{\bg}{\mathbf{g}}
\newcommand{\bp}{\mathbf{p}}

\newcommand{\bu}{\mathbf{u}}

\newcommand{\by}{\mathbf{y}}
\newcommand{\bz}{\mathbf{z}}

\newcommand{\cN}{\mathcal{N}}

\newcommand{\cX}{\mathcal{X}}
\newcommand{\cE}{\mathcal{E}}

\newcommand{\bbR}{\mathbb{R}}

\DeclareMathOperator*{\argmax}{arg\,max}

%% file: tex/0-abstract.tex
\begin{abstract}
Probabilistic state-estimation approaches offer a principled foundation for designing localization systems, because they naturally integrate sequences of imperfect motion and exteroceptive sensor data. Recently, probabilistic localization systems utilizing appearance-invariant visual place recognition (VPR) methods as the primary exteroceptive sensor have demonstrated state-of-the-art performance in the presence of substantial appearance change. However, existing systems 1) do not fully utilize odometry data within the motion models, and 2) are unable to handle route deviations, due to the assumption that query traverses exactly repeat the mapping traverse. To address these shortcomings, we present a new probabilistic \textit{topometric} localization system which incorporates full \mbox{3-dof} odometry into the motion model and furthermore, adds an ``off-map'' state within the state-estimation framework, allowing query traverses which feature significant route detours from the reference map to be successfully localized. We perform extensive evaluation on multiple query traverses from the Oxford RobotCar dataset exhibiting both significant appearance change and deviations from routes previously traversed. In particular, we evaluate performance on two practically relevant localization tasks: loop closure detection and global localization. Our approach achieves major performance improvements over both existing and improved state-of-the-art systems.

\end{abstract}
\begin{IEEEkeywords}
Localization, Autonomous Vehicle Navigation, Vision-Based Navigation
\end{IEEEkeywords}

%% file: tex/1-introduction.tex
\section{Introduction}

\begin{figure*}[t]
  \centering
  \includegraphics[width=0.93\linewidth]{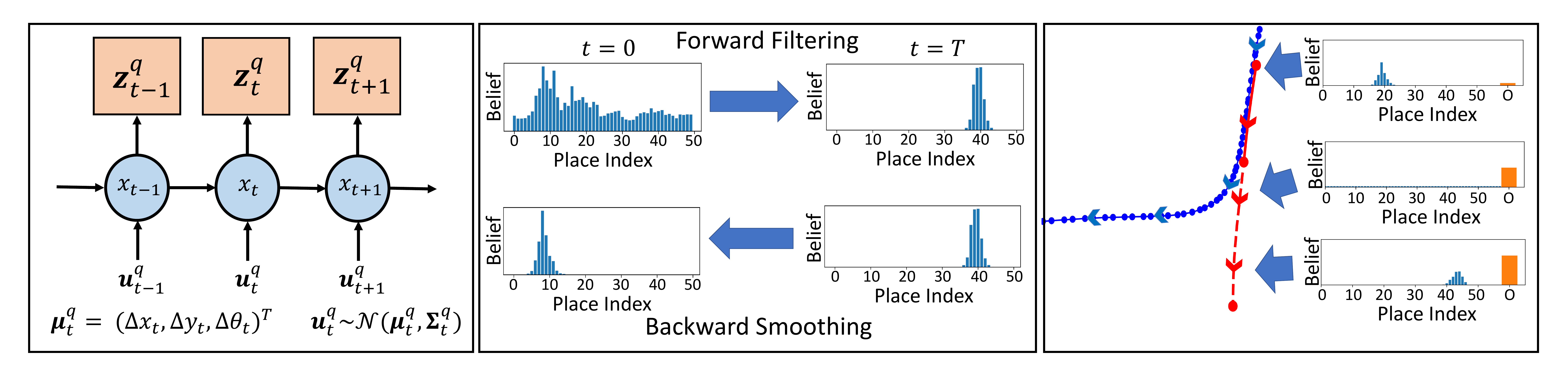}%
  \vspace{-0.3cm}
  \caption{We propose a probabilistic topometric localization system based on a discrete Bayes filter. \textbf{Left:} Bayes network diagram for our system, illustrating appearance-based measurements from query images $\bz_t^q$ and \mbox{3-dof} probabilistic query odometry $\bu_t^q$. \textbf{Middle:} We present methodology for loop closure detection combining our Bayes filter with backward smoothing, making data associations using a confidence score computed from the beliefs. \textbf{Right:} Our method includes a novel ``off-map" state, allowing the robot to infer whether or not it has left the map using appearance and odometry information.}%
  \label{fig:systemoverview}%
  \vspace{-0.3cm}
\end{figure*}

\IEEEPARstart{T}{his} letter addresses the long-term localization problem, where the aim is to localize a robot within a pre-built map across varying levels of appearance change induced by time of day, weather, seasonal or structural changes. We focus on topometric maps which do not require metrically accurate reconstructions of the environment, thus enabling large-scale mapping. Topological maps consist of a discrete set of nodes or ``places" connected by edges which encode spatial constraints. Incorporating relative pose information provided by odometry within these edges yields a \textit{topometric} representation. Each node within the map stores an appearance signature by applying appearance-invariant visual place recognition (VPR) techniques \cite{Torii15, triggs05, arandjelovic13, Arandjelovic16, revaud2019learning, cao2020unifying} to an image captured at the corresponding place, enabling persistent localization.

Localization systems built around appearance-based topometric maps have become a key component of the \mbox{6-dof} visual localization literature \cite{Sattler2018, Sarlin19, stenborg2020using}, where hierarchical localization approaches have demonstrated state-of-the-art performance. In such a hierarchy, the robot first localizes to a coarse, topometric map, and this coarse localization is used to initialize a structure-based pose refinement step. This hierarchical approach increases accuracy while minimizing computation time compared to only running pose refinement \cite{Sarlin19, stenborg2020using, xu2020probabilistic}.

Appearance-based topometric localization systems have also been used as the basis for mapping \mbox{systems~\cite{badino11, maddern12catslam, maddern12catgraph}}, allowing for large-scale mapping and navigation~\cite{maddern2012iros, furgale10, dayoub2013vision} with minimal compute and sensing capability (i.e.~camera and odometry). A key motivation for improving the capability of topometric localization systems is there to aid the development of downstream mapping and navigation systems.

Despite the current impressive performance of appearance-based localization systems, existing methods still have major capability gaps. Specifically, systems exploiting appearance-invariant VPR do not fully utilize \mbox{3-dof} odometry, suggesting possible further performance gains through its incorporation. Furthermore, existing systems tend to implicitly assume query traverses follow the same route as the mapping traverse. We show that localization performance of current systems is significantly compromised if this assumption is not met.

\vspace{0.05cm}
Our list of key contributions is the following:
\vspace{-0.005cm}
\begin{enumerate}
    \item A discrete filtering formulation for topometric localization which combines appearance-invariant VPR with \mbox{3-dof} odometry. Key to this system is a novel, principled technique for conditioning transition probabilities between \textit{discrete} states with \textit{continuous} \mbox{3-dof} odometry. This extends existing discrete filters which use none~\cite{Doan2019ICCV, xu2020probabilistic} to limited~\cite{stenborg2020using} odometry and provides an alternative to continuous state systems utilizing particle filters~\cite{xu2020probabilistic, doan2019visual}. We will show that discrete filters combined with backward smoothing provide substantially improved localization performance to continuous state systems which are only capable of forward filtering. 
    \item We propose a technique to handle localization of query traverses which can contain significant deviations in route compared to what was traversed when building the map. We achieve this by augmenting our discrete filter with an \textit{off-map state}. Importantly, addition of the off-map state does not hinder performance when the query traverse follows the reference traverse.
    \item Extensive evaluation of our system on both the loop closure detection problem and the global localization or ``wake-up" problem on the RobotCar dataset \cite{RobotCarDatasetIJRR}. We demonstrate that our improved model formulation yields significant performance gains for both tasks compared with the state-of-the-art~\cite{xu2020probabilistic, stenborg2020using}.
\end{enumerate}
Figure~\ref{fig:systemoverview} provides an overview of our proposed system. We also make our code freely available online\footnote{\githublink}.

%% file: tex/2-relatedworks.tex
\section{Related works}

This section reviews relevant research on visual place recognition (Section~\ref{subsec:RelVPR}) and appearance-based topometric localization systems (Section~\ref{subsec:RelAppearance}).

\subsection{Visual Place Recognition}
\label{subsec:RelVPR}

The VPR problem is commonly treated as a template matching problem~\cite{FischerGargIJCAI2021}; we assume the map consists of a set of reference ``places", and each place has an associated template extracted from either single images \cite{cummins08, Arandjelovic16, Torii15} or image sequences \cite{SeqSLAM, Hausler19, vysotska16}. Place recognition for a query template is achieved by retrieving the most ``relevant" template from the set of reference places. One focus within VPR research is around designing methods that can successfully match places across changing appearance conditions such as day/night and seasonal changes \cite{Torii15, Arandjelovic16, Sarlin19}, as well as across viewpoint changes \cite{cummins08, Garg18}. VPR approaches can be further partitioned into methods that perform matching using deep learning \cite{Arandjelovic16, revaud2019learning, Sarlin19,Hausler2021} (see~\cite{zhang2020visual} for a review) or handcrafted visual features \cite{Torii15, arandjelovic13} (see~\cite{lowry2016} for a review). Common across many of these approaches is a match scoring system; given a pair of images, a typical VPR system outputs a ``quality score" which indicates the likelihood of two images being captured from the same place. Methods such as \cite{Arandjelovic16, Torii15, revaud2019learning, Sarlin19} first embed the images into an embedding space and set the quality score as the Euclidean distance between the embeddings; where lower distances indicate a higher likelihood of a match. %

\subsection{Appearance-based Topometric Localization}
\label{subsec:RelAppearance}

One of the pioneering works on appearance-based localization is the probabilistic template matching approach FAB-MAP~\cite{cummins08}. FAB-MAP represents images using a bag-of-visual-words representation constructed from handcrafted SURF features, which severely limits its robustness to appearance changes. CAT-SLAM~\cite{maddern12catslam} augments FAB-MAP with local odometry information and uses a particle filter for localization. This trajectory-based algorithm was then reformulated as a graph-based representation in CAT-Graph~\cite{maddern12catgraph}. A key advantage of these approaches is that new place detection is formulated as a structured prediction task using the probabilistic model. Our method provides this capability with added appearance invariance through appearance-invariant VPR. Recently, Oliveira et al.~\cite{oliveira2020topometric} have proposed a probabilistic deep learning approach that consists of three modules: odometry estimation given two consecutive images using a Siamese network, a topological module that estimates the pose of an query image given the reference images, and a fusion module to merge the other two module outputs into a topometric pose. 

Xu et al.~\cite{xu2020probabilistic} adapt appearance-invariant image descriptors like NetVLAD~\cite{Arandjelovic16}, DenseVLAD~\cite{Torii15} and APGeM~\cite{revaud2019learning} to the Bayesian state-estimation framework, demonstrating state-of-the-art performance for global localization with substantial appearance change between query and reference images. The authors provide both a discrete topological filter that does not use odometry information, and a particle filter that utilizes odometry. While the particle filter uses odometry and demonstrates improved performance over the topological filter, it requires places in the map to be embedded within a global coordinate frame. Our method does not require this, making it more broadly applicable including situations where globally consistent reference poses are not available.

Badino et al.~\cite{badino11} proposed a topometric localization system, utilizing SURF features and odometry in a discrete Bayes filter formulation. Stenborg et al.~\cite{stenborg2020using} recently extended Badino's method to use NetVLAD~\cite{Arandjelovic16}, DenseVLAD~\cite{Torii15} and APGeM~\cite{revaud2019learning}. However, both methods utilize only 1D (forward) odometry information instead of full \mbox{3-dof} odometry utilized by our proposed method. Like our proposed approach, \cite{stenborg2020using} performs backward smoothing after forward filtering, however their method assumes no route deviations between query and reference. We show that this assumption has a substantial impact on localization performance when it does not hold. %

%% file: tex/3-methods.tex
\section{Methodology}

We now introduce the methodology behind our proposed localization system, starting with the map representation (Section~\ref{ssec:maprepresentation}) and state-estimation formulation for localization (Section~\ref{ssec:discretebayes}). We then introduce the motion and measurement models (Sections~\ref{ssec:motion} and~\ref{ssec:measurement}). This is followed by our proposed methodology for both the loop closure detection (Section \ref{ssec:lcdbackwards}) and global localization task (Section \ref{ssec:wakeup}). In the following section we use the superscript $r$ and $q$ to denote variables related to the reference map and query traverse, respectively. Vectors are printed in bold, with the $i$-th element of vector $\mathbf{v}$ being $v_i$.

\subsection{Map Representation}
\label{ssec:maprepresentation}

\begin{figure}[t]
  \centering
  \includegraphics[width=0.9\linewidth]{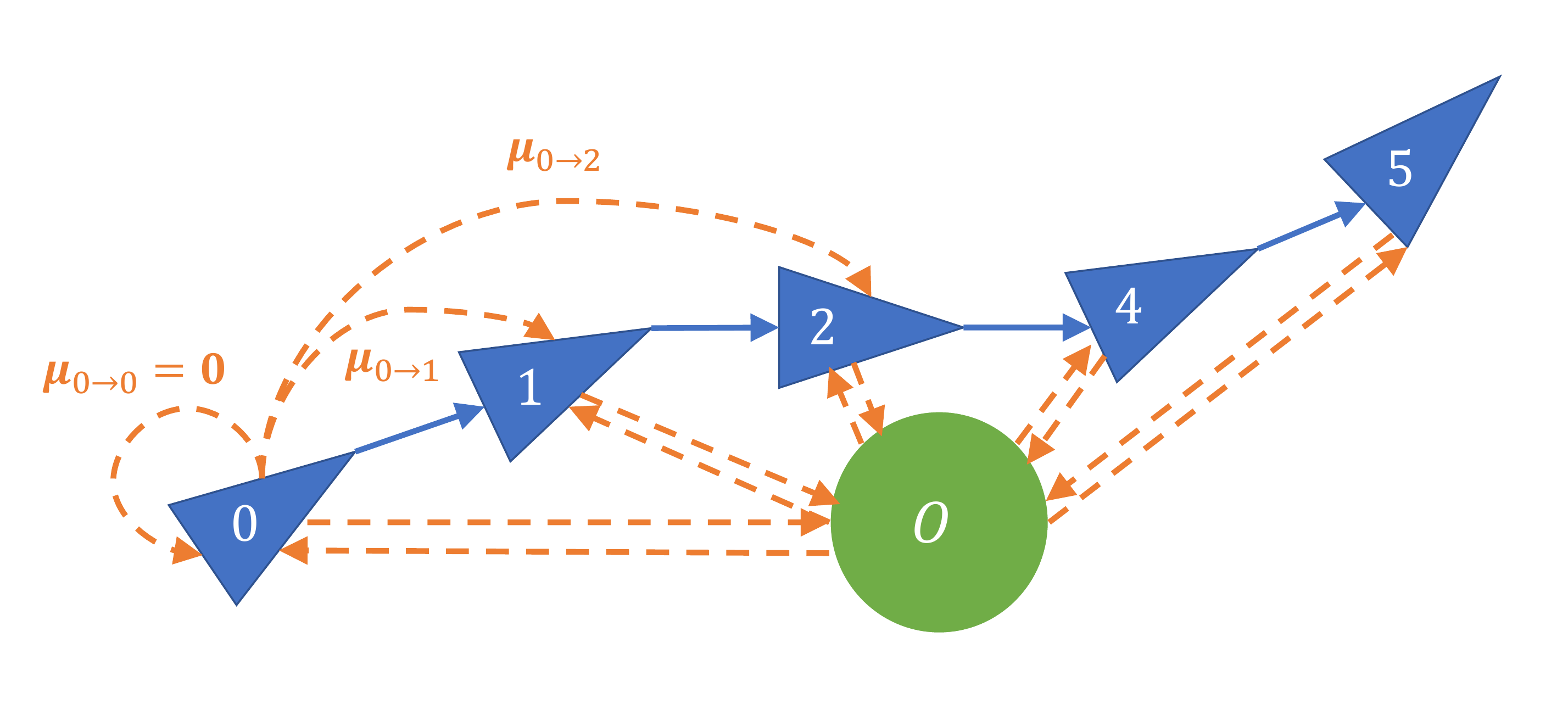}%
  \vspace{-0.4cm}
  \caption{Topometric map representation. Blue triangles indicate places within the map. Blue arrows represent adjacent places with odometry collected during mapping. Orange arrows indicate an edge with corresponding relative pose estimates. We also allow for transitions to/from the off-map node O.}%
  \label{fig:map_representation}%
  \vspace{-0.4cm}
\end{figure}

We assume a topometric representation for the reference map comprised of nodes and directed edges, with nodes $\cX = \{1, \dots, N\}$ indexed by integers representing places and edges $e_{i\rightarrow j}\in\cE$, for some $i, j\in \cX$ representing possible transitions between nodes. Each node $v\in\cX$ has an associated appearance signature $\bz_v^r\in \bbR^d$ given by an image embedding from a VPR technique such as \cite{Arandjelovic16, Torii15, revaud2019learning, Sarlin19} and each edge has an associated relative pose estimate $\bmu_{i\rightarrow j}^r = [\Delta x, \Delta y, \Delta \theta]^\top$ from odometry.

Nodes corresponding to nearby places are connected with edges. We assume a reference map is comprised of a single traverse, where node indices are ordered temporally, edges can be determined by index proximity, i.e. $e_{i\rightarrow j} \in \cE$ if $j > i$ and $j-i < w$ for some $w > 1$. In future work, we plan to generalize beyond this simple representation to an arbitrary network topology incorporating loop closures. In our experiments, we generate maps by subsampling nodes at regular spatial intervals as indicated by odometry.

Localization of a query image to a reference map involves identifying the reference node which corresponds to the same place where the query image was captured. If reference nodes are provided with associated poses (e.g.~from GPS), a metric pose estimate for a query image can be obtained by inheriting the pose of the selected reference node.

\subsection{Discrete Bayes Filters for Localization}\label{ssec:discretebayes}

We utilize a discrete Bayes filtering approach for our topometric localization system. For a query sequence comprised of appearance signatures $\{\bz_t^q\}_{t=0}^T$ and odometry $\{\bu_t^q\}_{t=1}^{T}$, the discrete Bayes filtering framework allows us to maintain a state estimate of the true unknown robot state $x_t$ at each timestep $t\leq T$. Under a discrete model, the true robot state takes on discrete values, in this case $x_t\in \cX \cup O$, where $O$ is a single ``off-map state" representing the case when the robot is not close to the previously traversed area given by the reference map. The state estimate for $x_t$ is given by a belief vector $\bp_t$, which is a probability density over discrete states. Given an initial belief $\bp_0$, we can update the belief recursively using a motion model (Section \ref{ssec:motion}) and measurement model (Section \ref{ssec:measurement}) which utilize the odometry and appearance information, respectively. Our proposed filtering system can be used to solve practical downstream localization tasks; in this letter we address both the loop closure detection (Section \ref{ssec:lcdbackwards}) and global localization (Section \ref{ssec:wakeup}) task. %
We now describe in detail our motion and measurement models.

\subsection{Motion Model}\label{ssec:motion}

Our motion model encodes the probabilistic dynamics between robot states and is represented by an $(N + 1) \times (N + 1)$ state-transition matrix $\bE_t$, where the row $i$ column $j$ element is defined as $E_{t, ij} = p(x_t = j | x_{t-1} = i, \bu_t^q)$. Unlike existing approaches, our discrete transition probabilities are conditioned on 3-dof query odometry and fully utilizes probabilistic motion models. Specifically, query odometry $\bu_t^q = (\bmu_t^q , \bSigma_t^q)$ contains a distribution over the change in pose from $\bz^q_{t-1}$ and $\bz^q_t$ parameterized by a mean vector $\bmu_t^q=(\mu^q_{x}, \mu^q_y, \mu^q_\theta)^\top$ and $3 \times 3$ covariance matrix $\bSigma_t^q$. This distribution, combined with the mean relative pose estimates encoded within map edges described in Section \ref{ssec:maprepresentation}, are used to compute transition probabilities between all nodes, including the off-map node $O$. %

\begin{figure}[t]
  \centering
  \includegraphics[width=0.95\linewidth]{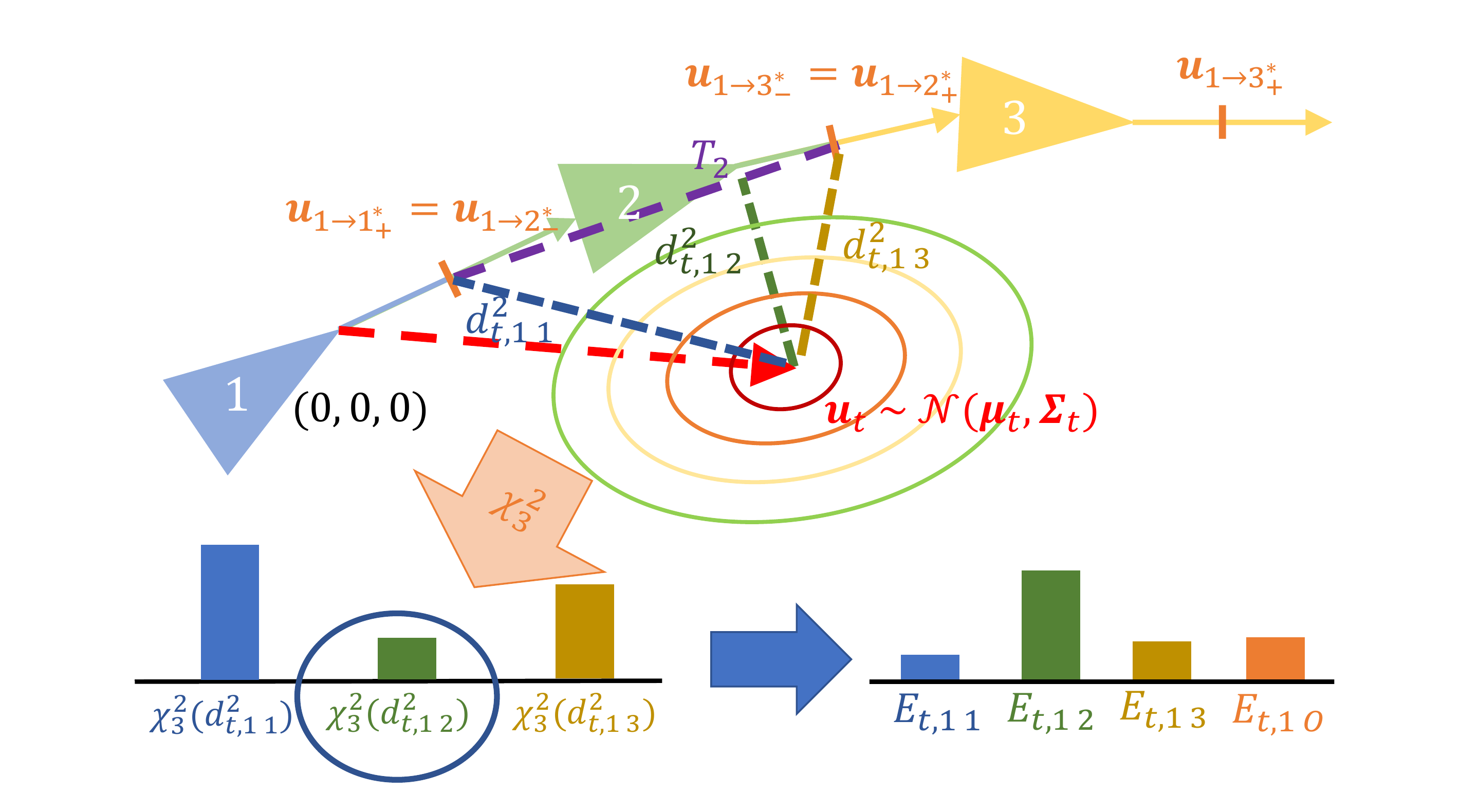}%
  \vspace{-0.3cm}
  \caption{Computation for transition probabilities for node 1. The local trajectory around nodes 1-3 (only $T_2$ in purple is shown) is scored against the probabilistic query odometry (red) using Mahalanobis distance to yield $d^2_{t, 1j}$. This is used for the off-map $E_{t, 1O}$ and within-map transitions $E_{t, 1j}$.}%
  \label{fig:motionupdate}%
  \vspace{-0.3cm}
\end{figure}

\textbf{Within-to-Within Map Transitions}\hspace{0.2cm} The transition probability between two within map nodes $i$ and $j$ denoted $E_{t, ij} = p(x_t = j | x_{t-1} = i, \bu_t^q)$, where $i, j \in \cX$ and edge $e_{i\rightarrow j}$ exists, is determined by the similarity in motion between the query odometry $\bu^q_t$ and the edge from $i$ to $j$. However, in the case where the query odometry is inconsistent to the outgoing edges from node $i$ (e.g.~query odometry measures 1.5m but neighboring reference nodes are spaced 1m apart), naively computing query and edge similarity will result in a low similarity for all outgoing transitions, even if the query odometry is consistent with the reference traverse. These discretization errors motivate the following methodology (see Figure~\ref{fig:motionupdate} for a visualization).

Let node $i$ be the origin of the local coordinate frame. The \mbox{3-dof} robot pose relative from node $i$ denoted $\by_t$ after applying query odometry $\bu_t^q$ is distributed as
\begin{equation}\label{eq:trueposedist}
    p(\by_t | x_{t-1} = i, \bu_t^q) \sim \cN(\bmu_t^q, \bSigma_t^q).
\end{equation}
This is a continuous distribution of relative pose $\by_t$ conditioned on discrete node $x_{t-1}=i$. We score the likelihood of $\by_t$ being on the previously mapped trajectory around node $j$ to evaluate $E_{t, ij}$. Denoting this trajectory segment as $T_j$, we approximate the true trajectory by interpolating the midpoints of the predecessor and successor node. Concretely,
\begin{align}\label{eq:nhoodtraj}
    T_j = \{s\, \bmu_{i\rightarrow j^*_+}^r +& (1-s) \bmu_{i\rightarrow j^*_-}^r,\: s\in[0, 1]\},
\end{align}
for midpoints $\bmu_{i\rightarrow j^*_{+}}^r = \frac{1}{2} \left( \bmu_{i\rightarrow j}^r + \bmu_{i\rightarrow j+1}^r \right)$ and $\bmu_{i\rightarrow j^*_{-}}^r = \frac{1}{2} \left( \bmu_{i\rightarrow j-1}^r + \bmu_{i\rightarrow j}^r \right)$. We set the transition probability using the likelihood at the most likely point in $T_j$ given by
\begin{align}\label{eq:maxlhood}
    E_{t, ij} \propto \max_{\by_t \in T_j} &p(\by_t | x_{t-1}, \bu_t^q).
\end{align}
To solve this, we reformulate this maximum likelihood problem as a minimum Mahalanobis distance problem due to the Gaussianity of $\by_t$, equivalently solving instead for
\begin{equation}\label{eq:minMNdist}
    d_{t, ij}^2 = \min_{s\in [0, 1]} \big(T_j(s) - \bmu_t^q\big)^\top \bSigma^{q-1}_t \big(T_j(s) - \bmu_t^q\big),
\end{equation}
where $T_j(s) = s\, \bmu_{i\rightarrow j^*_+}^r + (1-s) \bmu_{i\rightarrow j^*_-}^r$ for $s\in[0,1]$. The transition probability is now given by
\begin{equation}\label{eq:transition}
    E_{t, ij} = \frac{ e^{-\frac{1}{2} d^2_{ij} } }{ \sum_{j'\neq O} e^{-\frac{1}{2} d^2_{i j'}}} (1 - p_{t, iO}),
\end{equation}
where $E_{t, iO}$ is introduced below.

\textbf{Within-to-Off Map Transitions}\hspace{0.2cm} We can use the set of squared Mahalanobis distances $\{d_{t, ij}^2\}_{j\neq O}$ computed using \eqref{eq:minMNdist} to estimate the off-map transition probability $p_{t, iO}$. Intuitively, higher Mahalanobis distances imply a lower likelihood of the robot maintaining the original trajectory at the section, warranting a higher off-map transition probability. To make this relationship explicit, note that
\begin{equation}
    P(d_{t, ij}^2 < d) = \chi_3^2(d),
\end{equation}
where $\chi_3^2$ is the cumulative distribution function of a chi-squared distribution with 3 degrees-of-freedom. This represents the probability mass under the Gaussian in the hyperellipsoid defined by a maximum squared Mahalanobis distance of $d$. We set the off-map transition probability as
\begin{equation}\label{eq:chisqoffprob}
    E_{t, iO} = \min_j \chi_3^2(d^2_{ij}).
\end{equation}
For nodes where the query odometry is inconsistent to the mapping odometry, probability mass moves to the off-map state after a motion update step, reducing the total belief in the outgoing nodes. This is similar to particle re-weighting for odometry described in CAT-SLAM \cite{maddern12catslam} and CAT-Graph \cite{maddern12catgraph}. Figure~\ref{fig:motionupdate} illustrates our proposed motion model.

\textbf{Transitions From Off-Map}\hspace{0.2cm} The off-map node self-transition probability $E_{O\rightarrow O}$ is a parameter assumed constant for all $t$ and can be tuned on a representative training dataset. Increasing $E_{O\rightarrow O}$ increases the ``inertia" of the off-map state, making recovery to within-map states slower using the motion model alone. Furthermore, outward transitions from $O$ to any within-map state occur with uniform probability. 

\subsection{Measurement Model}\label{ssec:measurement}

Our measurement model at time $t$ is encoded by a length $N+1$ measurement vector $\bg_t$. The measurement model scores the likelihood of a query image given a particular robot state $p(\bz_t^q | x_t)$. For within-map states, we follow the method proposed in \cite{xu2020probabilistic}, where the likelihood for state $x_t$ is given by an exponential kernel over the appearance signatures
\begin{align}\label{eq:withinmeaslhood}
    p(\bz_t^q | x_t) \propto g_{t, x_t} = \exp(-\lambda \|\bz_t^q - \bz_{x_t}^r\|_2),
\end{align}
where $\lambda$ is calibrated using $\bz_0^q$ as in \cite{xu2020probabilistic}. %

The off-map likelihood is set as the likelihood of the $k^\text{th}$ best match for $k >> 1$. Concretely, $p(\bz_t^q | x_t = O) \propto g_{t, (k)}$, where $(k)$ is the index to the $k^\text{th}$ highest likelihood value at time $t$. The intuition behind this measurement update is to provide the off-map state a consistently high but not optimal likelihood. As a consequence, our method only converges to a location within the map if VPR yields consistently strong matches consistent with odometry. We now describe how we can use our localization system for two practical localization tasks: loop closure detection and global localization.

\subsection{Backward Smoothing for Loop Closure Detection}\label{ssec:lcdbackwards}

The loop closure detection (LCD) task involves localizing each element of a query sequence relative to the reference map. In this section, we will describe our proposed methodology for LCD combining discrete filters (Section \ref{ssec:discretebayes}) with the forward-backward algorithm for discrete hidden Markov models (Section \ref{ssec:lcdbackwards}). We note that while backward smoothing with a discrete filter has been investigated before in the localization context \cite{stenborg2020using}\footnote{Results from smoothing are used to remove unlikely image retrievals for a subsequent pose refinement step in a 6-dof localization pipeline.}, we are the first to present methodology for a standalone localization system. We will also show in Section \ref{ssec:resultsloop} that backward smoothing with discrete filters outperforms continuous state formulations where only forward filtering is possible.

The forward-backward algorithm allows us to recover a \textit{smoothed} belief $\bp_t^s$ at each time given \textit{future and past} odometry and appearance information. Specifically,
\begin{equation}
    \bp_{t, i}^s = p(x_t = i | \bu^q_{1:T}, \bz^q_{0:T}) \quad \text{where } t\leq T.
\end{equation}
In practice we observe that incorporating future observations in estimating the belief increases localization performance significantly compared to forward filtering alone at minimal computation cost. To apply the forward-backward algorithm requires three steps: extracting measurement vectors $\{\bg_t\}$ and state transition probabilities $\{\bE_t\}$, running forward recursion and finally running backward recursion.

Forward recursion computes for all $t$ the joint likelihood of observations up to time $t$ given by $\balpha_t$, starting with \mbox{$\balpha_0 = \bp_0 \circ \bg_0$} ($\circ$ is the elementwise product). We update $\balpha_t$ recursively using
\begin{equation}\label{eq:fwrecursion}
    \alpha_{t, i} = p(\bz^q_{0:t}, \bu^q_{1:t}, x_t=i) = g_{t, i} \sum_{j\in \cX} \alpha_{t-1, j} E_{t, ji}.
\end{equation}
This recursion given by \eqref{eq:fwrecursion} is fast to compute in practice due to the typical sparsity of edges in the map representation.

Backward recursion computes for all $t$ the joint likelihood of observations after time $t$ given by $\bbeta_t$, starting with $\bbeta_T = \mathbf{1}$. We update $\beta_t$ recursively using
\begin{equation}\label{eq:bwrecursion}
    \beta_{t-1, i} = p(\bz_{t:T}^q, \bu^q_{t:T}, x_t = i) = \sum_{j\in \cX} \beta_{t, j} g_{t, j} E_{t, ij}.
\end{equation}
We can finally recover the smoothed belief using the relation $\bp_t^s \propto \balpha_t \circ \bbeta_t$. More detail around the forward-backward algorithm can be found in~\cite{bishop2006prml}. After computing the smoothed belief, we can apply convergence detection described in Section \ref{ssec:onlineloc} to infer precise location estimates. Algorithm \ref{alg:fwbw} outlines the full procedure for the LCD task.

\begin{algorithm}[t]\label{alg:fwbw}
\caption{Loop closure detection}
\SetAlgoLined
\KwData{Query sequence, reference map and $\tau_\text{thres}$}
\KwResult{Sequence of state estimates $\{\hat{x}_t\}_{t=0}^T$ associated with each query image.}
 Initialize $\bp_0$ with uniform belief over within-map states and off-map state $p_{0, O}$\;
 Evaluate measurement model, yielding $\bg_t$ for all $t$\;
 Evaluate motion model, yielding $\bE_t$ for all $t$\;
 Run forward recursion, yielding $\balpha_t$ for all $t$\;
 Run backward recursion, yielding $\bbeta_t$ for all $t$\;
 Compute smoothed beliefs $\bp_t^s$ for all $t$\;
 Apply convergence detection to smoothed beliefs\;
\end{algorithm}

\vspace*{-0.3cm}
\subsection{Global Localization (Wakeup) Problem} \label{ssec:wakeup}

In addition to LCD, we can apply our localization system to the global localization or ``wakeup" task. Global localization involves localizing a robot to a map assuming no prior information is available around its location. Xu et al.~\cite{xu2020probabilistic} proposed a method for addressing the wakeup task using probabilistic filtering techniques, where the robot is first initialized with a uniform prior belief. Forward filtering steps are then applied given a query sequence until the belief has sufficiently converged around a specific place within the map; see~\cite{xu2020probabilistic} for detailed methodology. We show the utility of our filtering formulation for the wakeup task in our experiments.

\subsection{Convergence Detection}\label{ssec:onlineloc}

We use a modified version of the method proposed in \cite{xu2020probabilistic} to infer a single robot state $\hat{x}_t$ given a belief $\bp_t$. Convergence occurs when the belief is sufficiently concentrated around its mode, representing confidence in the state estimate. Denote the belief restricted to within-map nodes by $\bp_{t, \lnot O}$. Convergence score $\tau_t=s(\bp_t)\in[0, 1]$ is given by
\begin{equation}\label{eq:confidence}
    \tau_t = \sum_{x\in \cN_r(\hat{x}_t)} \bp_{t, x}, \:\: \text{where} \: \hat{x_t} = \argmax \bp_{t, \lnot O},
\end{equation}
where $\cN_r(\hat{x}_t)$ contains all nodes within radius $r$ around $\hat{x}_t$. If $\tau_t > \tau_\text{thres}$, set $\hat{x_t}$ to be the proposed robot location otherwise infer that the robot is off-map.

%% file: tex/4-experiments.tex
\section{Experimental Setup}

We evaluate the localization performance of our method against existing state-of-the-art and show both the robustness of our approach across varying levels of appearance change and changes in route between query and reference. 

\subsection{Datasets}

We use one reference and four particularly challenging query traverses from the Oxford RobotCar dataset~\cite{RobotCarDatasetIJRR} in our experiments, utilizing images from the forward facing camera and provided visual odometry (VO) data\footnote{The visual odometry (VO) in Oxford RobotCar is based on~\cite[Chapter~2]{churchill2012experience}. Our method is not constrained to using VO; any odometry source that provides relative pose estimates modeled by a Gaussian would be suitable.}. 

\begin{figure}[t]
  \centering
  \includegraphics[width=0.70\linewidth]{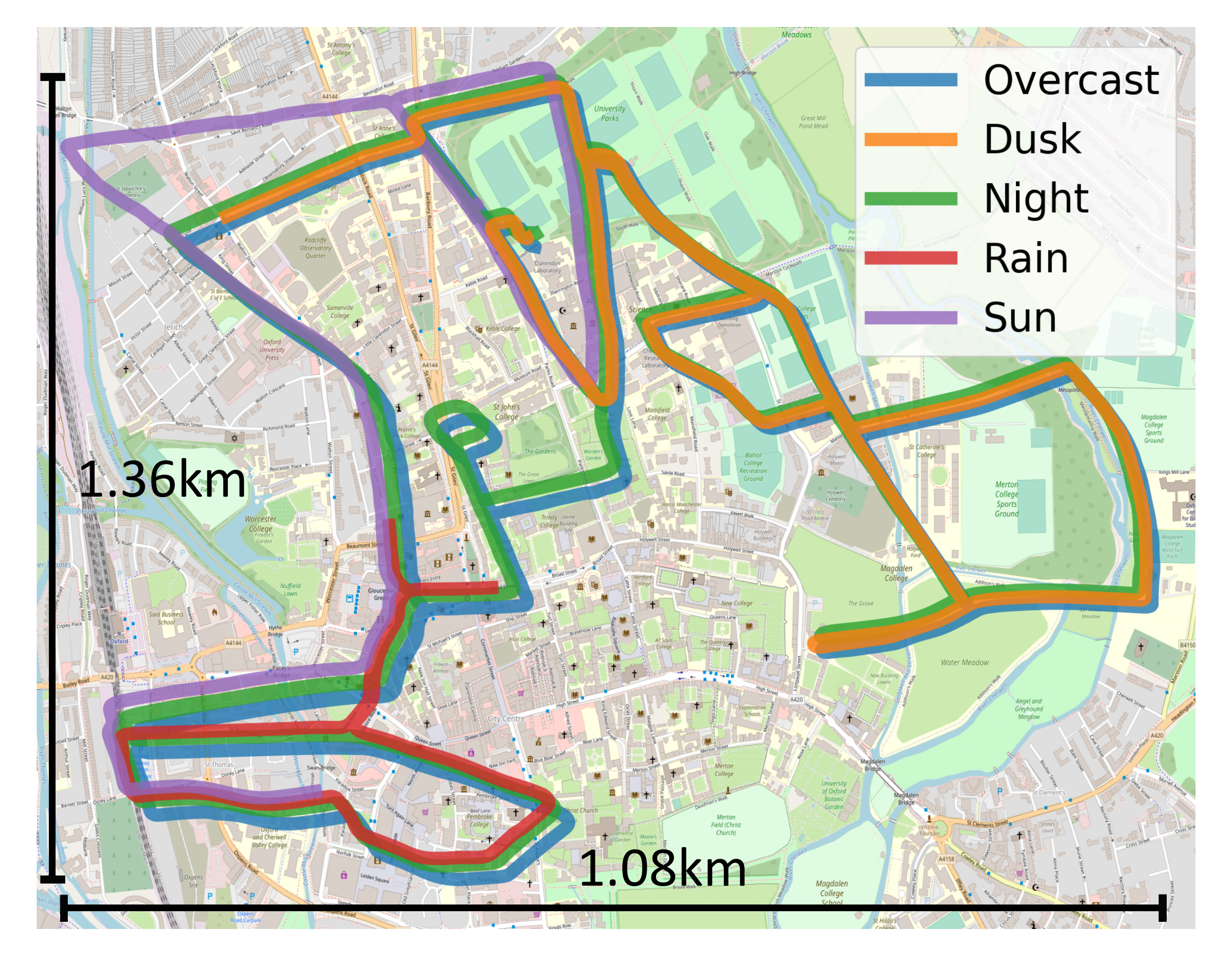}%
  \vspace{-0.3cm}
  \caption{Map overlaid with reference and query traverses used for evaluation. Credit: ©OpenStreetMap contributors}%
  \label{fig:robotcarmap}%
  \vspace{-0.4cm}
\end{figure}

The selected traverses exhibit substantial time-of-day, seasonal and structural changes between reference and all queries. Table \ref{tab:datasetrobotcar} shows summary statistics of the different traverses and Figure \ref{fig:robotcarmap} illustrates the selected routes. We note that the reference, dusk query and night query traverses are consistent with the ones used in Xu et al.~\cite{xu2020probabilistic}.

\begin{table}[t]
\centering
\caption{Oxford RobotCar Dataset Summary}

\begin{tabular}{|c | c c c c|} 
 \hline
 Seq & Date + Time & Dist. (km) & Conditions & Detour \\
 \hline
 Ref & 2015/03/17 11:08:44 & 9.01 & Overcast & N/A \\
 1 & 2014/11/21 16:07:03 & 6.18 & Dusk & NO \\
 2 & 2014/12/16 18:44:24 & 9.07 & Night & NO \\
 3 & 2015/07/29 13:09:26 & 1.86 & Rain & YES \\ 
 4 & 2015/04/24 8:15:07 & 3.46 & Sun & YES \\
 \hline
\end{tabular}
\label{tab:datasetrobotcar}
\end{table}

Images are captured at approximately 0.5m and 3m intervals, consistent with~\cite{xu2020probabilistic} for reference and query images, respectively. However, \cite{xu2020probabilistic} determined image spacing according to RTK GPS ground truth whereas we use imperfect VO in the spirit of building topometric maps without accurate reference pose information. RTK GPS ground truth~\cite{RobotCarRTKarXiv} is used for evaluation purposes only, to measure the pose error between proposed reference and query image. A query image is defined as being off-map if there are no reference map images within 5m and 30 deg based on the ground truth.%

\subsection{Comparison Methods}

We compare our proposed method against the following comparison methods and ablations:

\begin{enumerate}
    \item Single image VPR, where the best matching reference image is used to localize (without using odometry or filtering, denoted as ``Baseline'' in Figures~\ref{fig:loops}-\ref{fig:wakeup}).
    \item Topological Bayes filter from Xu et al.~\cite{xu2020probabilistic}, which does not use odometry information.
    \item Topometric localization from Stenborg et al.~\cite{stenborg2020using}, which uses forward from odometry motion only.
    \item Monte Carlo Localization (MCL) method from Xu et al.~\cite{xu2020probabilistic}. This requires RTK-GPS ground truth poses for map nodes, which is not the case for our proposed system. We run MCL five times and present the best set of results out of the five trials.
    \item Our proposed method removing the off-map state, denoted ``No Off" in Figures~\ref{fig:loops}-\ref{fig:wakeup}, utilizing \mbox{3-dof} odometry for within-map nodes transition probabilities.
\end{enumerate}
All methods utilize our proposed convergence detection and backward smoothing is applied for (2-3) for a fair evaluation of the state estimation formulation for both localization tasks.

\subsection{Sensor Setup}

We utilize two VPR methods in our experiments, a weaker descriptor HF-Net~\cite{Sarlin19} optimized for mobile platforms~\cite{li2020dxslam} as well as a more powerful but resource intensive descriptor NetVLAD~\cite{Arandjelovic16} extracted on full resolution images. For odometry, we use the raw \mbox{6-dof} relative poses provided by stereo visual odometry, projected to \mbox{3-dof}. The odometry motion model as described in~\cite{Thrun2005} was used to provide odometry mean and covariances. Odometry is of especially poor quality on the night and dusk query traverses. %

\subsection{Parameters}

For each VPR method used, we maintain a single set of parameters for all traverses and both tasks. Parameters are optimized on the LCD task, providing the best possible performance attainable by each system. Specific parameter values are provided in the available open-source code.

%% file: tex/5-results.tex
\section{Results}\label{sec:results}

We measure the performance of our proposed system against the comparison methods on the loop closure detection (Section \ref{ssec:lcdbackwards}) and global localization (Section \ref{ssec:wakeup}) tasks. We demonstrate the utility of our 3-dof motion model and off-map state on the RobotCar dataset. In addition, we provide timing information for all steps detailed in the methodology.

\subsection{Loop Closure Detection}\label{ssec:resultsloop}

\begin{figure*}[ht]
  \centering
  \includegraphics[width=0.99\linewidth]{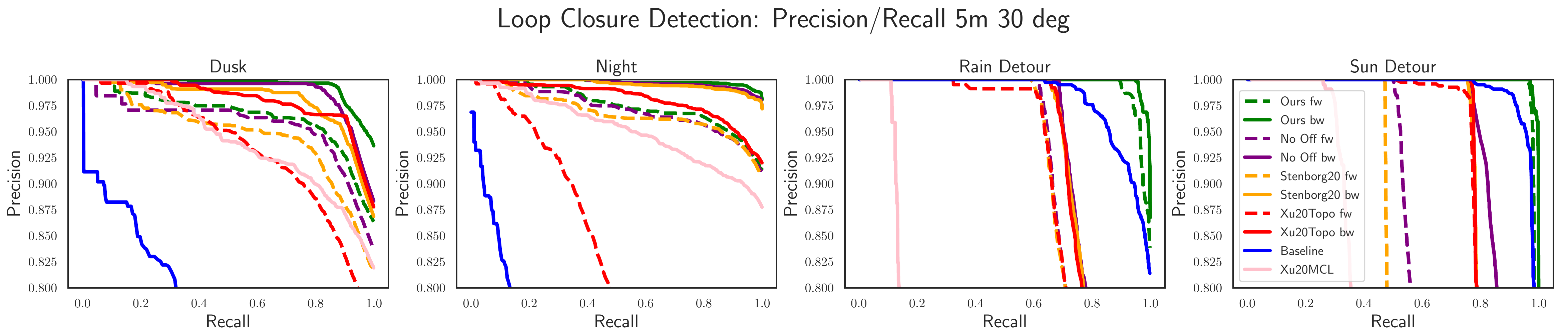}%
  \caption{Precision/Recall curves for the loop closure detection task. Dashed lines represent detected loop closures using forward filtering only, solid lines include backward smoothing for discrete filtering methods. Our proposed motion model consistently yields higher PR curves compared to using forward odometry only (Stenborg20). Discrete filters with backward smoothing also outperforms the continuous-state approach using MCL (Xu20MCL). For the rain and sun traverses, our proposed method greatly outperforms the comparison methods due to the inclusion of the off-map state.}
  \label{fig:loops}%
  \vspace{-0.1cm}
\end{figure*}

\begin{figure*}[ht]
  \centering
  \includegraphics[width=0.99\linewidth]{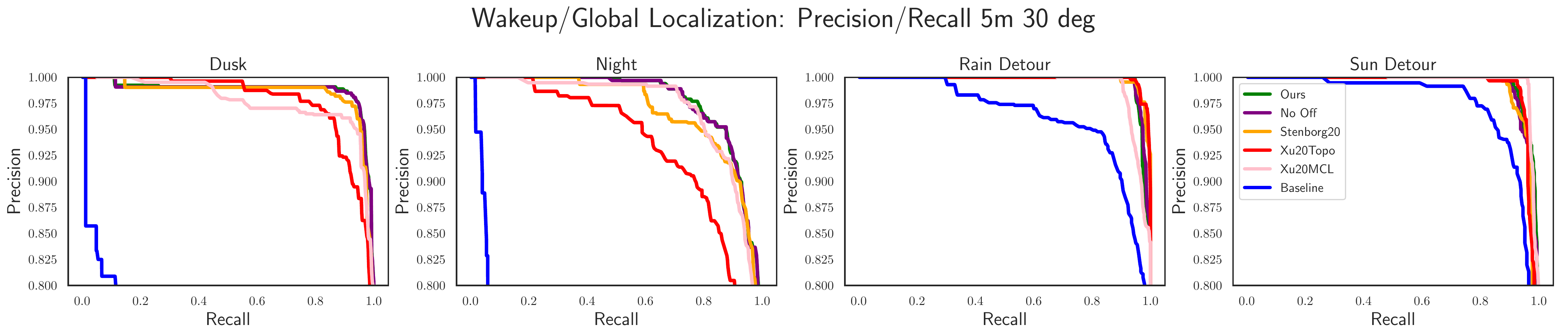}%
  \caption{Precision/Recall curves for the global localization task. Our proposed discrete filter yields comparable performance to the continuous-state MCL approach. The incorporation of the off-map state does not hinder localization performance on dusk and night where no route deviations are present.}
  \label{fig:wakeup}
  \vspace{-0.1cm}
  \end{figure*}

For the loop closure detection (LCD) task, we evaluate our method using standard precision/recall (PR), noting that for the rain and sun traverses there are sequence elements which can yield true negatives for loop proposals due to the robot being off-map. PR curves for the HF-Net~\cite{Sarlin19} descriptor are presented in Figure~\ref{fig:loops}, generated by varying the threshold $\tau_\text{thres}$ or the image similarity score for the single image baseline. As an ablation, we also show results where no backward smoothing is applied, only forward filtering. In addition, Table \ref{tab:loopsummary} summarizes the maximum recall at the 99\% precision level (R@99\%P) for all methods and provides a comparison to the more powerful (but slower) NetVLAD descriptor~\cite{Arandjelovic16}. We observe that using NetVLAD~\cite{Arandjelovic16} yields appreciable performance gains over HF-Net~\cite{Sarlin19} for methods (2) and (3) for the challenging night traverse as well as the single image VPR baseline for the rain and sun traverses. However, for our proposed method, the utility of using a more powerful VPR yields minimal performance gains overall.

A clear motivation for using a discrete filtering formulation for localization as opposed to a continuous state approach (e.g. MCL) is the availability of backward smoothing. For the dusk query traverse with HF-Net descriptor, forward filtering alone for our method results in a 9\% R@99\%P which is lower than the 23\% yielded by the Xu20MCL method~\cite{xu2020probabilistic}. Upon application of backward smoothing, our method increases to 94\% R@99\%P, which is a vast improvement above MCL.

In addition, changing from 1-dof odometry (Stenborg20~\cite{stenborg2020using}) to 3-dof odometry (Ours) yields appreciable performance gains across all traverse, up to 39\% absolute increase in R@99\%P on the night traverse for HF-Net. The performance gains are typically more appreciable when odometry is degraded relative to the reference map, which is the case for challenging query conditions (i.e. night and dusk).

\begin{figure}[t]
  \centering
  \includegraphics[width=0.73\linewidth]{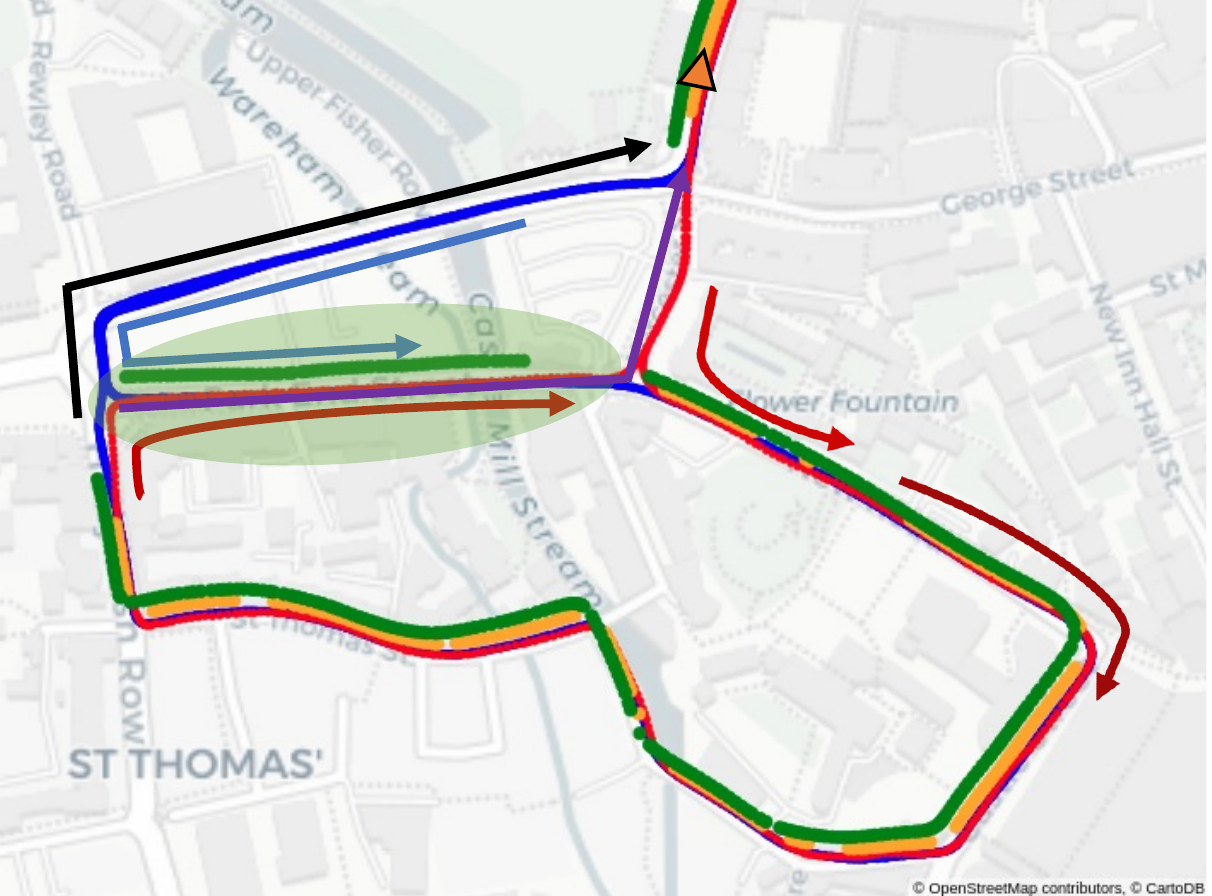}%
  \vspace*{-0.2cm}
  \caption{Localization results for the rain query traverse. Reference map (blue), query sequence (red), LCD predictions from our approach (green) and Stenborg20~\cite{stenborg2020using} (orange). Arrows indicate direction of travel with both reference and query moving clockwise around the loop. Our approach handles all route deviations whereas Stenborg20 fails in the area covered by the green ellipse. Interestingly, Stenborg20 successfully regains tracking near the end of the sequence (orange arrow) since the final detour (purple) is similar in distance to the equivalent mapping segment (black).} %
  \label{fig:qualitative}%
  \vspace{-0.5cm}
\end{figure}

The rain and sun query traverses contain major route deviations from the reference map, where the robot deviates from the reference map and rejoins at a later point. These traverses illustrate the limitations of existing discrete filtering formulations, which assume possible state transitions from a given node correspond only to spatially proximal nodes. In our experiments, this translates to a limit on R@99\%P, with discrete filtering approaches that do not incorporate an off-map state yielding 77\% and 76\% R@99\%P for the rain and sun traverses respectively. By incorporating the off-map state to our discrete filtering formulation however, R@99\%P increases from 69\% (No Off) to 96\% (Ours) for the rain query and from 76\% to 98\% for the sun query. Figure~\ref{fig:qualitative} illustrates this qualitatively for the rain traverse.

Finally, we note that including the off-map state when the query traverse follows the same route as the reference map (e.g.~dusk and night queries) does not appear to degrade localization performance, yielding a decrease only on the night traverse (HF-Net) of 10\% R@99\%P.

\begin{table}[t!]
\centering
\caption{Recall at 99\% Precision, Loop Closure Detection at 5m, 30 deg (HF-Net/NetVLAD)}
\begin{tabular}{| c | c | c | c | c |} 
 \hline
 & Dusk & Night & Rain & Sun \\
 \hline
 Baseline & 0.01 / 0.00 & 0.01 / 0.00 & 0.60 / 0.76  & 0.38 / 0.85  \\ 
 \hline
 Ours & \textbf{0.92} / \textbf{0.88}  & 0.86 / \textbf{0.96}  & \textbf{0.97} / \textbf{0.96}  & \textbf{0.96} / \textbf{0.98}  \\ 
 \hline
  No Off & 0.86 / 0.82  & \textbf{0.96} / 0.94 & 0.69 / 0.69  & 0.75 / 0.76  \\ 
 \hline
 Stenborg20 & 0.57 / 0.64  & 0.57 / 0.89 & 0.56 / 0.66  & 0.37 / 0.76  \\ 
 \hline
  Xu20Topo & 0.47 / 0.44  & 0.41 / 0.49  & 0.19 / 0.77 & 0.20 / 0.67 \\ 
 \hline
 Xu20MCL & 0.23 / 0.22   & 0.49 / 0.18   & 0.11 / 0.28  & 0.29 / 0.29  \\ 
 \hline
\end{tabular}
\vspace{-0.4cm}
\label{tab:loopsummary}
\end{table}

\subsection{Global Localization}\label{ssec:resultsglb}

We evaluate our system on the global localization or ``wake-up" task by running 500 trials with initialization points randomly sampled\footnote{The same initialization points are used for all comparison methods and ablations to ensure a fair comparison.} along the full query traverse, following the configuration in Xu et al.~\cite{xu2020probabilistic}. We apply forward filtering using the query sequence until either convergence occurs, or 30 steps have been taken ($\approx 90$m traveled). For the single image baseline, we take the position of the best match of the first image from each query sequence in a given trial. We measure localization performance by precision/recall, with PR curves presented in Figure \ref{fig:wakeup} and summary statistics (R@99\%P) presented in Table \ref{tab:wakeupsummary}. We furthermore present the mean distance traveled until convergence for each method in Table \ref{tab:wakeupsteps}. Similar to the LCD experiments, we provide results for both HF-Net~\cite{Sarlin19} and NetVLAD~\cite{Arandjelovic16}.

Consistent with the results presented in the LCD task, the incorporation of the off-map state in our discrete filtering formulation does not hinder localization performance when no route detours are present, with a maximum difference of 4\% R@99\%P across dusk and night. In addition, we note that our proposed discrete filtering formulation with 3-dof odometry yields comparable (occasionally surpassing) performance to the MCL method which utilizes additional information in the form of RTK pose estimates for the reference images. The mean distance traveled for our discrete filter at the 99\% precision level is similar to MCL, meaning the similar localization performance is not at the expense of localization latency. Consistent with LCD experiments, incorporation of more odometry information yields better localization performance compared to 1-dof (Stenborg20~\cite{stenborg2020using}) or no (Xu20Topo~\cite{xu2020probabilistic}) odometry.

\begin{table}[t!]
\centering
\caption{Recall at 99\% Precision, Global Localization at 5m, 30 deg (HF-Net/NetVLAD)}
\begin{tabular}{| c | c | c | c | c |} 
 \hline
 & Dusk & Night & Rain & Sun \\
 \hline
 Baseline & 0.01 / 0.01  & 0.02 / 0.01  & 0.33 / 0.74 & 0.75 / 0.79  \\ 
 \hline
 Ours & \textbf{0.87} / 0.69  & \textbf{0.69} / 0.78  & 0.95 / 0.96 & 0.92 / \textbf{0.98}  \\ 
 \hline
 No Off & 0.87 / 0.65  & 0.67 / \textbf{0.79}  & 0.95 / 0.95 & 0.90 / 0.97  \\ 
 \hline
 Stenborg20 & 0.83 / \textbf{0.89}  & 0.59 / 0.63 & 0.95 / \textbf{0.98} & 0.90 / 0.97  \\ 
 \hline
  Xu20Topo & 0.55 / 0.70 & 0.21 / 0.38 & 0.96 / 0.98 & \textbf{0.94} / \textbf{0.98} \\
 \hline
 Xu20MCL & 0.45 / 0.80 & 0.70 / 0.68 & \textbf{0.97} / 0.94 & 0.91 / 0.97 \\ 
 \hline
\end{tabular}
\label{tab:wakeupsummary}
\end{table}

\begin{table}[t!]
\centering
\caption{Mean distance traveled (m) until convergence at 99\% Precision and 5m, 30 deg (HF-Net/NetVLAD)}
\begin{tabular}{| c | c | c | c | c |} 
 \hline
 & Dusk & Night & Rain & Sun \\
 \hline
 Baseline & 0.0 / 0.0  & 0.0 / 0.0  & 0.0 / 0.0 & 0.0 / 0.0  \\ 
 \hline
 Ours & 45.8 / 49.8 & 51.5 / 50.2 & 25.8 / 19.2 & 24.7 / 17.2 \\ 
 \hline
  No Off & 45.8 / 50.6  & 51.6 / 49.3  & 27.2 / 20.3 & 25.7 / 16.2  \\ 
 \hline
 Stenborg20 & 51.8 / 40.2 & 59.2 / 58.9 & 35.1 / 19.4 & 27.9 / 17.1 \\ 
 \hline
  Xu20Topo & 47.5 / 38.9 & 50.0 / 54.0 & 24.9 / 14.3 & 20.1 / 15.0  \\ 
 \hline
 Xu20MCL & 52.2 / 43.0 & 50.6 / 52.9 & 23.3 / 13.6 & 24.2 / 10.8 \\ 
 \hline
\end{tabular}
\label{tab:wakeupsteps}
\end{table}

\subsection{Timings}

We provide detailed timing information for the required steps assuming feature extraction is performed by HF-Net~\cite{Sarlin19}. All methodology was implemented in Python and benchmarked on a desktop with a Intel® Core™ i7-7700K CPU, 32Gb RAM running Ubuntu 20.04 with no GPU. Feature extraction and evaluating the motion model accounts for the majority of computation time, with backward smoothing requiring minimal additional computational overhead. The total computational time is around 100ms, which allows processing 10 steps per second. As query images are taken at 3m intervals, this allows real-time processing of images captured by a robot traveling up to 30m/s (110km/h).

\begin{table}[t!]
\centering
\caption{Detailed Timings (ms) per iteration}
\vspace*{-0.5cm}
\begin{tabular}[t]{| c | c | c | c | c |}
 \hline
 Feat.~extr.~& Motion & Meas.~& Forward & Backward\\
 \hline
 52 & 30 & 9 & 5 & 3 \\
 \hline
\end{tabular}
\label{tab:timings}
\end{table}

%% file: tex/6-conclusion.tex
\section{Discussion and Conclusion}

In this work we presented a new topometric localization system based on discrete Bayes filters which utilizes full \mbox{3-dof} odometry information for localization as well as new place awareness. We demonstrated state-of-the-art localization performance on two localization tasks, with substantial improvements in the case where the query traverse deviates from the map. Future work will generalize our system to handle arbitrary network topologies within the reference map. In addition, we aim to extend our localization system into a multi-session mapping system, leveraging the strong appearance invariance in VPR for long-term mapping.